\documentclass[preprint,12pt]{elsarticle}

\usepackage{hyperref,tikz,bbding,amssymb,pifont,adjustbox,graphicx,soul,textcomp,array,longtable,bm,multirow,xcolor,color,hyphenat}

\graphicspath{ {./figures/} }

\usepackage[left=1in, right=1in, top=1in]{geometry}

\usepackage[T1]{fontenc}
\usepackage{makecell}

\usepackage{etoolbox}
\makeatletter
\def\ps@pprintTitle{%
 \let\@oddhead\@empty
 \let\@evenhead\@empty
 \def\@oddfoot{}%
 \let\@evenfoot\@oddfoot}
\makeatother

\newcommand{\cmark}{\ding{51}}%

\newcolumntype{P}[1]{>{\centering\arraybackslash}p{#1}}
\newcolumntype{M}[1]{>{\centering\arraybackslash}m{#1}}

\begin{document}

\begin{frontmatter}

\title{Understanding Spatial Language in Radiology: Representation Framework, Annotation, and Spatial Relation Extraction from Chest X-ray Reports using Deep Learning}

\author[mymainaddress]{Surabhi Datta}
\ead{surabhi.datta@uth.tmc.edu}
\author[mymainaddress]{Yuqi Si}
\author[mysecondaryaddress]{Laritza Rodriguez}
\author[mysecondaryaddress]{Sonya E Shooshan}
\author[mysecondaryaddress]{Dina Demner-Fushman}
\author[mymainaddress]{Kirk Roberts}
\ead{kirk.roberts@uth.tmc.edu}


\address[mymainaddress]{School of Biomedical Informatics, The University of Texas Health Science Center, Houston, TX}
\address[mysecondaryaddress]{National Library of Medicine, National Institutes of Health, Bethesda, MD
}

\begin{abstract}
We define a representation framework for extracting spatial information from radiology reports (Rad-SpRL). We annotated a total of 2000 chest X-ray reports with 4 spatial roles corresponding to the common radiology entities. Our focus is on extracting detailed information of a radiologist's interpretation containing a radiographic finding, its anatomical location, corresponding probable diagnoses, as well as associated hedging terms. For this, we propose a deep learning-based natural language processing (NLP) method involving both word and character-level encodings. Specifically, we utilize a bidirectional long short-term memory (Bi-LSTM) conditional random field (CRF) model for extracting the spatial roles. The model achieved average F1 measures of $ 90.28 $ and $ 94.61 $ for extracting the \textsc{Trajector} and \textsc{Landmark} roles respectively whereas the performance was moderate for \textsc{Diagnosis} and \textsc{Hedge} roles with average F1 of $ 71.47 $ and $ 73.27 $ respectively. The corpus will soon be made available upon request.
\end{abstract}

\begin{keyword}
\texttt Spatial Relations\sep Radiology Report\sep NLP \sep Deep Learning
\end{keyword}

\end{frontmatter}

\section{Introduction}

There has been a growing interest in automatically extracting useful information from unstructured reports in the medical domain. Radiology reports contain a wealth of important information and have been one of the most explored free text clinical report types for information extraction using NLP. Automatic recognition of clinically important information such as actionable findings and their corresponding location and diagnoses facilitates the time-consuming process of manual review of the reports containing radiologists' descriptions of imaging results. 


Besides radiology-specific knowledge and experience, the ability to mentally visualize complex 3D structures based on the interpretation of 2D images (e.g., X-ray diffraction images) remains vital for radiologists. A few studies \cite{Birchall2015spatial} \cite{Corry2011future} have highlighted the possible requirement of spatial ability skills in prospective radiologists to perceive and understand the spatial relationships between different objects in radiology practice. These spatial interpretations from images are also summarized in the corresponding free text reports. Thus, radiology report texts have a high prevalence of spatial relations in the way radiologists describe imaging findings and their association with anatomical structures. These spatial relations provide sufficient contextual information to visualize a radiologist's interpretations of the findings or imaging observations with reference to a body location. Moreover, some of these spatially-grounded findings demand immediate action by the referring physician. Therefore, it is beneficial to generate structured representations by understanding the spatial meanings from these unstructured reports, which can then be used for a variety of purposes such as easy review of the important actionable findings by the referring clinicians, analysis for predictive modeling, and automatically generating more complete annotations containing spatial and diagnosis-related information of findings for associated images.

In the general domain, earlier studies \cite{kordjamshidi2010spatial}\cite{Kordjamshidi2017spatial} have formulated and evaluated the spatial role labeling (SpRL) task for extracting spatial information from text by mapping language to a formal spatial representation. In the SpRL annotation scheme, \textit{an object of interest} (\textsc{Trajector}) is associated with \textit{a grounding location} (\textsc{Landmark}) through \textit{a preposition or spatial trigger} (\textsc{Spatial Indicator}). For example, in the sentence, \textit{``The book is on the table''}, the spatial preposition \textit{`on'} indicates the existence of a spatial relationship between the object \textit{`book'} (\textsc{Trajector}) and its location \textit{`table'} (\textsc{Landmark}). 

In the medical domain, one study so far by Roberts et al. \cite{Roberts2015consumer} has used the SpRL scheme to extract spatial relations between symptoms/disorders and anatomical structures from consumer-related texts. Similar spatial roles can be constructed for radiology text. For instance, in a radiology report sentence, \textit{``Mild streaky opacities are present in the left lung base''},
the location of a clinical finding \textit{`opacities'} (\textsc{Trajector}) has been described with respect to the anatomy \textit{`left lung base'} (\textsc{Landmark}) using the spatial preposition \textit{`on'} (\textsc{Spatial Indicator}). 

Moreover, radiologists oftentimes document potential diagnoses related to the clinical findings which are spatially grounded. Consider the following example: 

\begin{itemize}
  \item[] \textit{Stable peripheral right  lower lobe opacities seen \textbf{between} the anterior 7th and 8th right ribs  which may represent pleural reaction or small pulmonary nodules.}
\end{itemize}

\noindent Here, presence of a finding -- \textit{`stable peripheral right lower lobe opacities'} at a specific location -- \textit{`anterior 7th and 8th right ribs'} may elicit the radiologist to document two possible diagnoses -- \textit{`pleural reaction'} and \textit{`small pulmonary nodules'}. Since the actual occurrence of a disorder is highly dependent on various patient factors such as other physical examinations, laboratory tests, and symptoms, the radiologists usually describe diagnoses with uncertainty phrases or hedges. For instance, in the example above, the hedge term \textit{`may represent'} is used to relate a finding and its corresponding body location with the most probable diagnoses.

Our work recognizes granular information about the interpreted diagnoses by identifying them in context to the same spatial preposition (e.g., \textit{in}, \textit{of}, \textit{within}, \textit{around}) connecting a clinical finding to an anatomical location. Thus, we extract detailed information about a finding, the body location where the finding is detected, possible diagnoses associated with the finding, and also any hedging term used by radiologists in interpreting these diagnoses. Additionally, the finding and the location terms contain their respective descriptors (e.g., the descriptor \textit{`mild streaky'} associated with the finding \textit{`opacities'}).


In this paper, we propose a framework as a preliminary step to understand textual spatial semantics in chest X-ray reports. We define a basic spatial representation framework that extends SpRL for radiology (Rad-SpRL) involving interactions among common radiology entities.
Since most of the actionable clinical findings in all types of radiology reports are spatially located and represent a probable diagnosis, Rad-SpRL can be easily extended to other report types. Consider the following sentence from a head CT report: 

\begin{itemize}
  \item[] \textit{A well circumscribed hypodense 1 cm lesion is seen \textbf{in} the right cerebellar hemisphere consistent with prior stroke.}
\end{itemize}

\noindent Here, the spatial preposition \textit{`in'} describes that the finding \textit{`lesion'} is located inside the anatomical structure \textit{`right cerebellar hemisphere'} which is also consistent with the diagnosis \textit{`stroke'}. To evaluate this representation, we manually annotated a corpus of 2000 radiology reports (a subset of publicly available OpenI chest X-ray reports \cite{Demner-Fushman2016preparing}) using Rad-SpRL. Owing to the promising results of applying deep learning models in entity and relation extraction, we have adopted a bidirectional long short-term memory (Bi-LSTM) recurrent neural network with a conditional random field (CRF) layer as our baseline model to identify detailed spatial relationships, including diagnosis and hedging terms from the reports.


The organization of the rest of the paper is as follows. Section~\ref{sect:related_work} highlights the previous relevant studies on chest radiology, spatial relation extraction, and the OpenI dataset \cite{Demner-Fushman2016preparing}. Section~\ref{sect:dataset} describes our new corpus of 2000 chest X-ray reports annotated according to Rad-SpRL and the annotation process that produced this corpus. Section 4 includes a description of our baseline system based on Bi-LSTM-CRF for extracting fine-grained spatial information, including the implementation details. Results are summarized in Section 5, while Section 6 contains a discussion of the results. Section 7 concludes the paper and provides directions for future work.


\section{Related work}
\label{sect:related_work}
\subsection{Types of chest radiology entities extracted}
Numerous studies have focused on extracting specific information such as clinical findings or imaging observations, differential diagnoses, and anatomical locations from chest-related reports (highlighted in Table~\ref{table:chest_x-ray_studies}). In Table~\ref{table:chest_x-ray_studies}, we compare the specific information types or the radiology entities extracted in the previous studies from chest radiology reports using NLP. We primarily pay attention to the clinically-important entities which are common across various types of radiology reports. We also do not take into account the cases where uncertainty and negation information were used to detect the presence or absence of a particular finding or a disease \cite{Wang2017chestxray8}. For example, \textit{Hedge} is not considered as extracted in Table~\ref{table:chest_x-ray_studies} when the uncertainty levels are classified into negative, uncertain or positive for each finding term extracted \cite{Irvin2019chexpert}. Further, in Table~\ref{table:chest_x-ray_studies}, we have not considered studies dealing with specific body locations (e.g., mammography reports containing breast imaging information, and head CT reports) since the entities of interest are usually very domain-specific such as `Clock face', `Depth', `BI-RADS category' etc. in case of mammography reports. We also do not take into account the works which focused on detecting a specific disease such as pneumothorax \cite{Wang2019enhanceddiagnosis} or pulmonary lesion \cite{Pesce2019learning} from chest radiographs. Note that these works did not attempt to extract all the information types collectively, neither did they focus on identifying any association or relation among these entities.

\begin{table}[ht]
\caption{Comparison of our corpus with the studies extracting clinically-relevant radiological entities from chest X-ray and chest CT reports.}
\centering
\resizebox{\textwidth}{!}{
\begingroup
\renewcommand{\arraystretch}{1.2}
\begin{tabular}{lcccccccc}
  \hline
 \textbf{Paper} & \textbf{Finding} & \textbf{Anatomy} & \textbf{Descriptor} & \textbf{Diagnosis} & \textbf{Device} & \textbf{Hedge} & \textbf{Negation} & \textbf{Relation} \\ 
  \hline
  Hassanpour et al. \cite{Hassanpour2016information} & \cmark \ & \cmark & \cmark &  &  & \cmark & \cmark &  \\ 
  \hline
   Cornegruta et al. \cite{Cornegruta2016modelling} & \cmark & \cmark & \cmark &  & \cmark &  & \cmark &  \\ \hline
  Bustos et al. \cite{Bustos2019padchest} & \cmark & \cmark &  & \cmark &  &  &  &  \\
  \hline
  Hassanpour et al. \cite{Hassanpour2017characterization} & \cmark &  &  &  &  &  &  &  \\
  \hline
  Irvin et al. \cite{Irvin2019chexpert} & \cmark &  &  & \cmark & \cmark &  & \cmark &  \\
  \hline
  Annarumma et al. \cite{Annarumma2019automated} & \cmark & \cmark & \cmark &  &  &  &  &  \\
  \hline
  Wang et al. \cite{Wang2017chestxray8} & \cmark &  &  & \cmark &  &  &  &  \\
  \hline
  Rad-SpRL (this paper) & \cmark & \cmark & \cmark & \cmark &  & \cmark &  & \cmark \\
  \hline
\end{tabular}
\endgroup
}
\label{table:chest_x-ray_studies}
\end{table}

\subsection{Relation extraction including spatial relation}
In context to detecting relations among radiology entities, one study by Sevenster et al. \cite{Sevenster2012correlating} has built a reasoning engine to correlate clinical findings and body locations in radiology reports utilizing the Medical Language Extraction and Encoding System (MedLEE). However, the major limitation of this work is the system's poor recall. Yim et al. \cite{yim2016tumor} has worked on extracting relations containing tumor-specific information from radiology reports of hepatocellular carcinoma patients.
In Table~\ref{table:sp_rl_studies}, we present the two works relevant to spatial information extraction from radiology reports. The main limitations of Rink et al. \cite{Rink2013extracting} are the usage of appendicitis-specific lexicons and the requirement of manual effort in crafting rules based on syntactic dependency patterns to identify the spatially-grounded inflammation description. Besides being domain-specific, another limitation of Roberts et al. \cite{Roberts2012amachine} is that the study extracts only the location entities associated with an actionable finding and this required relying on heavy feature engineering.

\begin{table}[ht]
\caption{Studies focusing on spatial relations in radiology reports.}
\centering
\tabcolsep=0.06cm
\resizebox{\textwidth}{!}{
\begingroup
\renewcommand{\arraystretch}{1.2}
\begin{tabular}{lcccc}
  \hline
 \textbf{Paper} & \textbf{Finding} & \textbf{Anatomy} & \textbf{Diagnosis} & \textbf{Hedge} \\ 
  \hline
  
  Roberts et al. \cite{Roberts2012amachine} & --- & \makecell{\cmark \\ (Spatially related to a finding)} & --- & --- \\
  \hline
  Rink et al. \cite{Rink2013extracting} & \cmark & \makecell{\cmark \\ (Linked with finding)} & --- & \makecell{\cmark \\ (Not linked with finding/location)}  \\
  \hline
\end{tabular}
\endgroup
}
\label{table:sp_rl_studies}
\end{table}

\subsection{Studies using OpenI X-ray report annotations}
\begin{figure*}[t!]
\includegraphics[width=1.1\linewidth]{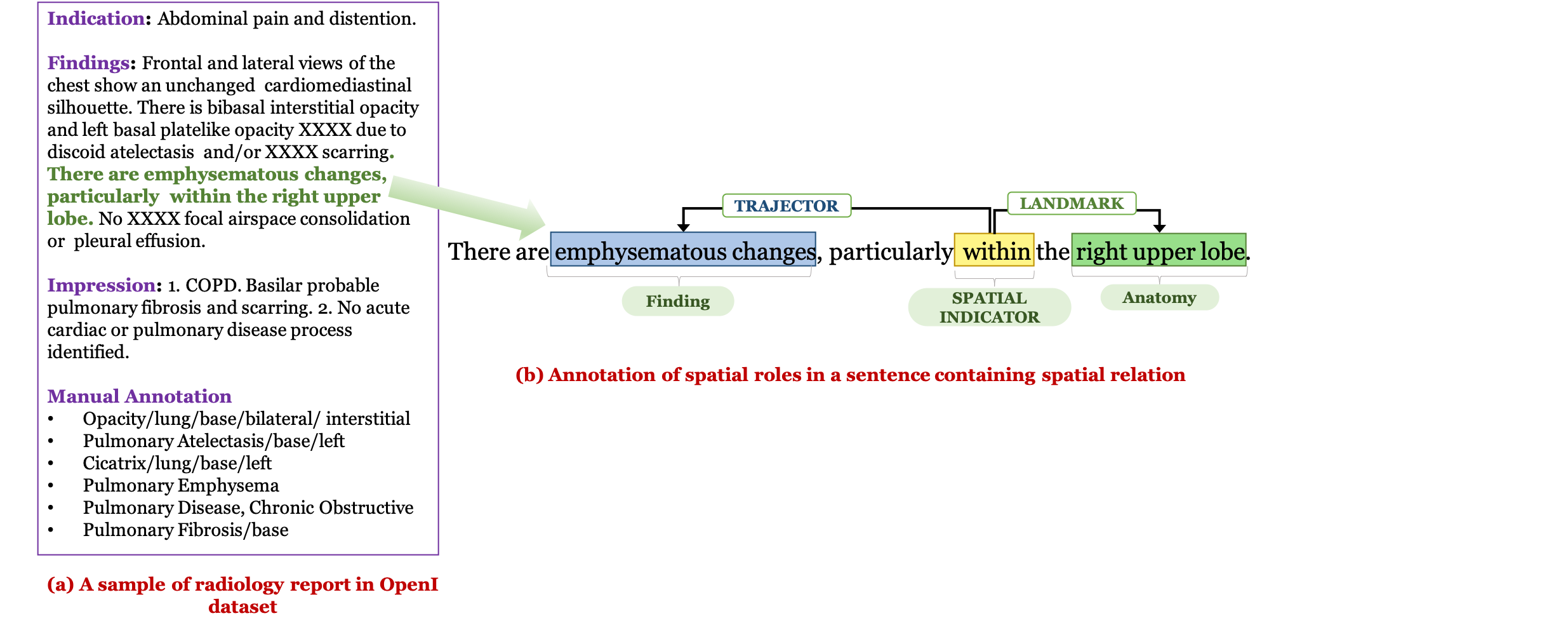}
\centering
\caption{Examples of manual annotations: (a) OpenI annotations, (b) Our spatial relation annotations.}
\label{fig:openi_example}
\end{figure*}

OpenI is a biomedical image search engine\footnote{https://openi.nlm.nih.gov/}. One of its datasets is a public chest X-ray dataset containing 3,955 de-identified radiology reports from the Indiana Network for Patient Care released by National Library of Medicine \cite{Demner-Fushman2016preparing}. 
(Hereafter referred to simply as the OpenI dataset.)
We have presented an example of the manual annotation of a sample report in the OpenI dataset in Figure~\ref{fig:openi_example} (the annotations are inspired by MeSH terms). Although most of the OpenI annotations embody the relationship between finding and location, there are, however, a few missing relations. For example, note that in Figure~\ref{fig:openi_example} the OpenI manual annotations contain the normalized finding \textit{Pulmonary Emphysema} corresponding to the phrase \textit{`emphysematous changes'} in the report, but do not annotate the associated location \textit{`right upper lobe'}. The OpenI dataset has been used previously in many studies, presented in Table~\ref{table:openi_dataset_studies}. However, most of the these studies focused on the extraction of only the disease/finding \cite{Wang2017chestxray8}, \cite{Wang2018tienet}, \cite{Peng2018negbio}, \cite{Daniels2019exploiting}, \cite{Zech2018variable}.
Two studies worked on automatically annotating both disease and disease descriptions (e.g., location, severity) \cite{Shin2016learning}, \cite{Huang2019anannotationmodel} similar to the human annotations in Demner-Fushman et al. \cite{Demner-Fushman2016preparing}. However, all these works ignored distinguishing diagnosis terms from findings (except for Peng et al. \cite{Peng2018negbio}), and annotating correlations between them. We describe annotation-specific limitations of each of these works in Table~\ref{table:openi_dataset_studies}.

\begin{longtable}{|p{0.1\linewidth}|p{0.4\linewidth}|p{0.4\linewidth}|}
\caption{\hbox{Studies who have used OpenI manual annotations.}} \label{table:openi_dataset_studies} \\
  \hline
 \textbf{Paper} & \textbf{How OpenI chest X-ray dataset is involved} & \textbf{Limitation (Radiology entities annotated/considered for model evaluation)}\\ 
  \hline
  Demner-Fushman et al. \cite{Demner-Fushman2016preparing} & Manually annotated or coded the collected 3996 reports with findings, diagnoses, body parts using MeSH terms supplemented by RadLex codes. Automatic annotation was also produced by the Medical Text Indexer (MTI). & This is a manual annotation process relying on MeSH terms and standard qualifier terms. The coded terms were not well-distinguished between findings and diagnoses. Moreover, the annotation lacks other information such as relation between findings and diagnoses. The automatic labeling does not include the related body parts for the labeled finding. (\textbf{Positive \hbox{Findings}/\hbox{Diagnoses} and Body parts}) \\
  \hline
  Shin et al. \cite{Shin2016learning} & Trained CNNs using already available image annotations from Demner-Fushman et al. \cite{Demner-Fushman2016preparing} and considered images where each image is labeled with only a single disease label using unique MeSH term combinations (this accounted for around 40 percent of the full OpenI dataset and 17 unique disease annotation patterns). 
  Generated image annotations including disease as well as its contexts such as location, severity, and the affected organs by taking into account image/text contexts while training CNNs. & Although the annotation includes disease context, it only generates image caption corresponding to one disease per image. However, it is highly likely that multiple diseases or findings may be associated with an image. (\textbf{Findings/Diagnoses and their context such as location and severity}) \\
  \hline
  Wang et al. (2017) \cite{Wang2017chestxray8} & Used text mining (DNorm \cite{leaman2015dnorm} and MetaMap \cite{Aronson2010metamap}) to label disease names using reports. Evaluated their image labeling method on OpenI reports using the key findings/disease names coded by human annotators as gold standard \cite{Demner-Fushman2016preparing}. & Only used the available annotations for evaluating their proposed method. (\textbf{Findings/Diagnoses})  \\
  \hline
  Wang et al. (2018) \cite{Wang2018tienet} & Evaluated a text-image embedding auto-annotation framework on the OpenI dataset using the key findings/disease names coded by human annotators as the gold standard \cite{Demner-Fushman2016preparing}. & Used the annotated OpenI dataset for evaluating proposed disease classification method for 14 diseases. (\textbf{Findings/Diagnoses})  \\
  \hline
  Peng et al. \cite{Peng2018negbio} & Defined rules utilizing universal dependency graphs to identify negation or uncertainty related to findings. Manually checked the annotations in OpenI and organized the findings into 14 domain-important and generic types of medical findings. & Used the OpenI dataset both for designing the patterns and testing. Although they mentioned that organizing the findings into fine-grained categories can facilitate in correlating findings with the diagnosis, the terms distinguished as diagnoses or body parts were not utilized in the study for showing any correlation. (\textbf{Findings}) \\
  \hline
  Daniels et al. \cite{Daniels2019exploiting} & Proposed a deep neural network that predicts one or more diagnoses given an image by jointly learning visual features and topics from report findings. & Used the OpenI dataset and their corresponding `findings' annotations both for fine-tuning and evaluating the model. (\textbf{Findings/Diagnoses}) \\
  \hline
  Huang et al. \cite{Huang2019anannotationmodel} & Proposed a neural sequence-to-sequence model by leveraging ``indication'' information of the report which includes annotating relationship between the positions where the finding term appears. They used the OpenI manual annotations as a reference annotation for evaluating the model. & Although this generated annotations for multiple diseases per image and also aimed to improve the results of Shin et al.\cite{Shin2016learning} in annotating disease along with context such as location and severity, they did not annotate other useful contexts including spatial information of the finding as well as the associated diagnosis. (\textbf{Findings/Diagnoses and their context such as location and severity}) \\
  \hline
  Zech et al. \cite{Zech2018variable} & To assess the generalizability of a deep learning model for screening pneumonia across 3 hospital systems. Used human-annotated pathology labels of OpenI dataset for testing. & Used OpenI only for evaluation. (\textbf{Findings/Diagnoses}) \\
  \hline
  Candemir et al. \cite{Candemir2018deeplearning} & Fine-tuned several deep CNN architectures to detect presence of cardiomegaly. Used OpenI dataset both for training and testing. & Manually annotated each OpenI image into one of the
following severity categories: borderline, mild, moderate, severe, and  non-classified using the corresponding reports having cardiomegaly. (\textbf{Findings, specifically cardiomegaly and their severity levels}) \\
  \hline

\end{longtable}

\section{Proposed spatial relation annotation framework}
\label{sect:dataset}

\subsection{Dataset for spatial relation annotation}
A subset of 2000 reports from a total of 2470 non-normal reports as judged by two human annotators in Demner-Fushman et al. \cite{Demner-Fushman2016preparing} was used to create our spatial relation corpus. This newly annotated chest X-ray corpus contains spatial relations between findings and body locations as well as the correlated probable diagnoses and the hedging terms used in qualifying the diagnoses. We have presented a simple comparison between the OpenI manual annotations and our spatial annotations of a sample report in Figure~\ref{fig:openi_example}. Note that we have not annotated other findings appearing in the report such as \textit{Opacity} and \textit{Pulmonary Fibrosis} as their corresponding body locations are not described through any spatial preposition.

\subsection{Rad-SpRL}
Our spatial representation framework (Rad-SpRL) consists of 4 spatial roles (\textsc{Trajector}, \textsc{Landmark}, \textsc{Hedge}, and \textsc{Diagnosis}) with respect to a \textsc{Spatial Indicator}. The spatial roles and the \textsc{Spatial Indicator} are defined as follows:
\begin{enumerate}
  \item \textsc{Spatial Indicator}: term (usually a preposition, e.g., \textit{in}, \textit{within}, \textit{at}, \textit{near}) that triggers a spatial relation
  \item \textsc{Trajector}: object (finding, anatomical location) whose spatial position is being described
  \item \textsc{Landmark}: location of the \textsc{Trajector} (may also be chained as a \textsc{Trajector} to another \textsc{Landmark})
  \item \textsc{Hedge}: phrase indicating uncertainty (e.g., \textit{could be}, \textit{may represent}), generally in reference to the \textsc{Diagnosis} and very rarely in the \textsc{Trajector}
  \item \textsc{Diagnosis}: disease/clinical condition the radiologist associated with the finding
\end{enumerate}

\begin{figure*}[t!]
\includegraphics[width=1.1\linewidth]{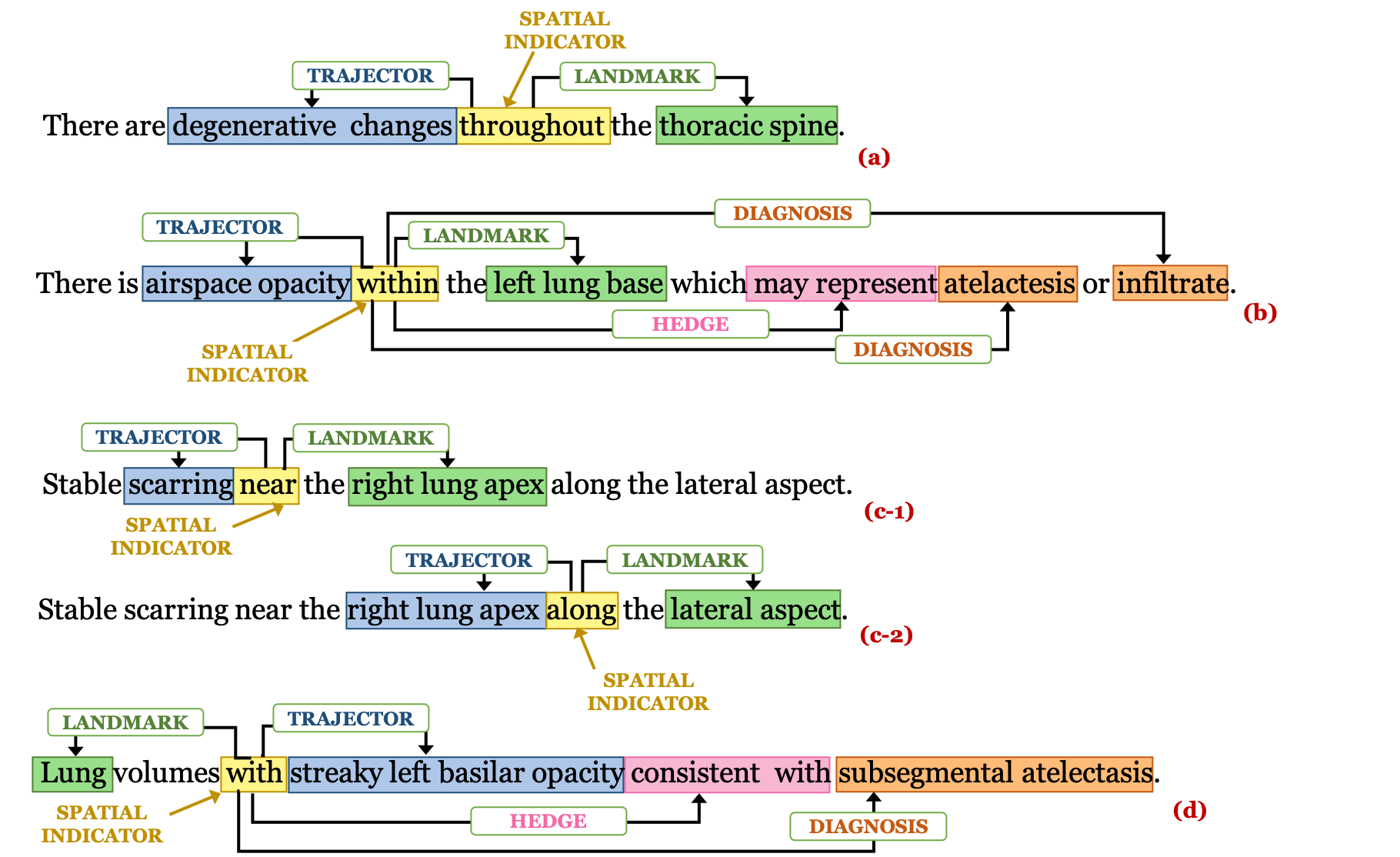}
\centering
\caption{Examples of spatial role annotations: (a)Sentence having \textsc{Trajector} and \textsc{Landmark}, (b) Sentence having \textsc{Trajector}, \textsc{Landmark}, \textsc{Hedge}, and \textsc{Diagnosis}, (c-1) and (c-2) show the annotations of the same sentence containing 2 \textsc{Spatial Indicator}s where the same entity \textit{right lung apex} acts as a \textsc{Landmark} in (c-1) and a \textsc{Trajector} in (c-2), and (d) Sentence where a \textsc{Landmark} is described with a \textsc{Trajector}.}
\label{fig:examples}
\end{figure*}

In most of the cases where a sentence contains spatial information, a finding (\textsc{Trajector}) is usually detected at a particular body location (\textsc{Landmark}) where the \textsc{Trajector} term appears to the left of the \textsc{Spatial Indicator} and the \textsc{Landmark} to its right. However, there are instances where a spatial preposition describes the body location (\textsc{Landmark}) with its associated abnormality (\textsc{Trajector}) and the \textsc{Trajector} term appears to the right of the \textsc{Spatial Indicator} and \textsc{Landmark} to the left (refer to example in Figure~\ref{fig:examples}(d)). We have presented a few specific examples to highlight how various spatial roles and \textsc{Spatial Indicator}s are identified in sentences following the above definitions of Rad-SpRL in Figure~\ref{fig:examples}. Please note that we have considered disease/condition terms as \textsc{Diagnosis} only when they are documented in conjunction with any spatially-located finding, or in other words are entirely probable diagnoses inferred from the finding. Also note that there is some ambiguity between a finding and a diagnosis, such that the same phrase may appear as a \textsc{Diagnosis} in one relation while being a \textsc{Trajector} in another. Our purpose here is not to formally distinguish between a finding and a diagnosis, but rather to identify the spatial relationships in radiology reports where the \textsc{Trajector} is generally a finding (or artifact in the image) and the \textsc{Diagnosis} is generally a well-understood disease term.

\subsection{Annotation process}
\begin{figure*}[t!]
\includegraphics[width=0.85\linewidth]{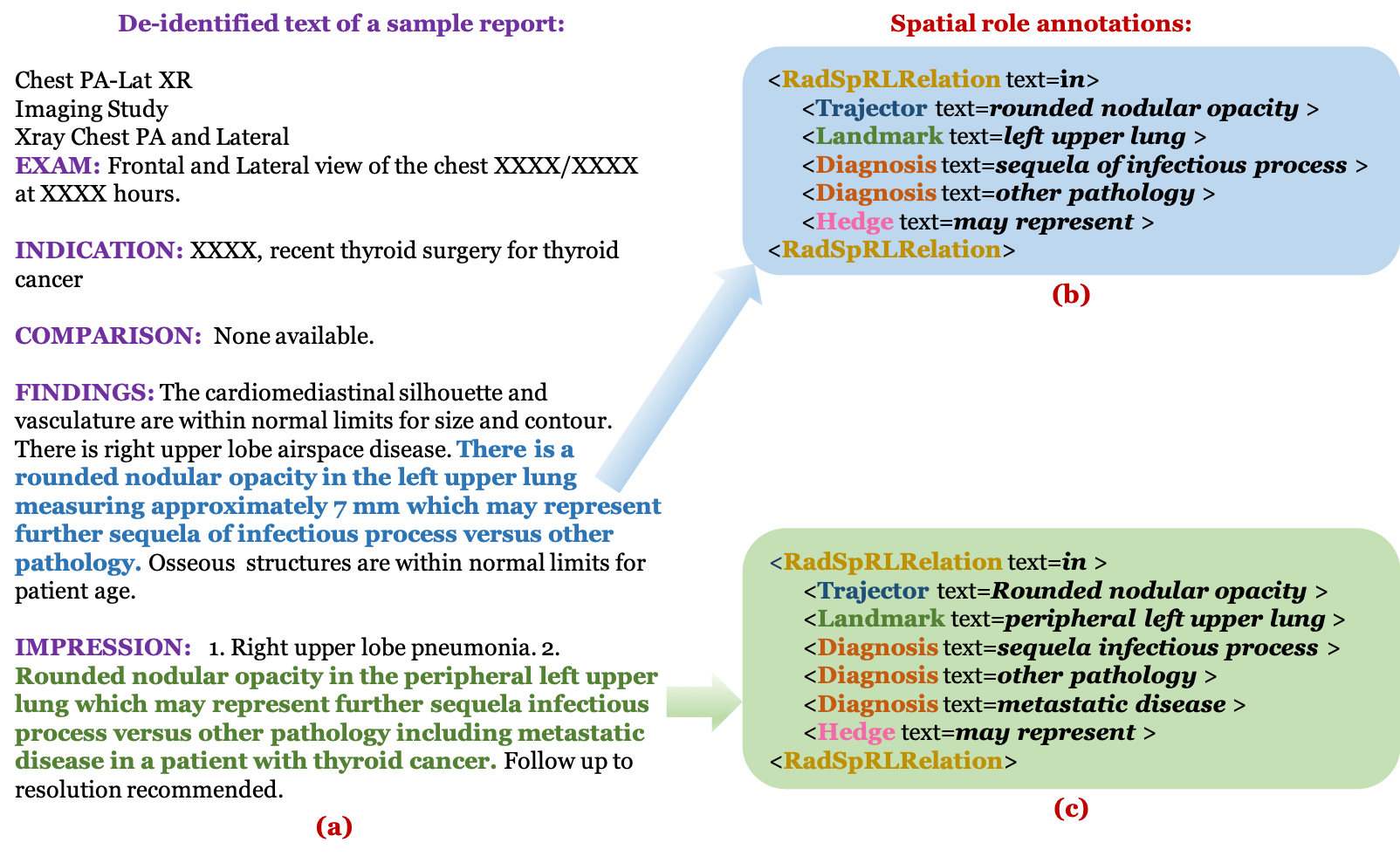}
\centering
\caption{(a) Example of a de-identified report in our corpus, (b) Spatial role label annotations for the sentence represented by blue text in (a), and (c) Spatial role label annotations for the sentence represented by green text in (a). RadSpRLRelation indicates the text of the respective \textsc{Spatial Indicator}s implying the existence of a spatial relation in both the sentences.}
\label{fig:report_annotation}
\end{figure*}

Two annotators (S.E.S., a medical librarian, and L.R., an MD) annotated the spatial roles for each identified  \textsc{Spatial Indicator} in each of the 2000 reports independently. They also were the annotators that manually coded the findings/diagnoses available as part of OpenI dataset \cite{Demner-Fushman2016preparing}. The spatial relation annotations were conducted in two rounds and reconciled after each. The first round consisted of annotating the first 500 reports and the second round consisted of annotating the remaining 1500. Figure ~\ref{fig:report_annotation} shows a sample annotated report from the corpus.

\subsubsection{Annotation agreement}
The inter-annotator agreement statistics for both \textsc{Spatial Indicator} and spatial roles are shown in Table~\ref{table:inter-annotator_agreement}. The Kappa ($\kappa$) agreement between the two annotators has been calculated for \textsc{Spatial Indicator} (as this is a binary classification task) whereas we report the overall F1 agreement for annotating the spatial role labels (as this is a role identification task). The Kappa agreement is high for \textsc{Spatial Indicator}s in both annotation rounds. The F1 agreements for the 4 spatial roles are fairly low in the first round with much improvement in the second round. This is mainly because it is relatively easy and unambiguous to locate a spatial preposition in a sentence compared to identifying the spatial roles. All conflicts were reconciled with an NLP expert (K.R.) following each round of annotation. The moderate agreement rate for \textsc{Trajector} and \textsc{Diagnosis} roles was likely due to ambiguity in distinguishing the two roles in a sentence, especially when the language pattern is different from the usual. Consider the examples below:
\begin{enumerate}
  \item \textit{Probably scarring \textbf{in} the left apex, although difficult to exclude a  cavitary lesion.}
  \item \textit{There are irregular opacities \textbf{in} the left lung apex, that could represent  a cavitary lesion \textbf{in} the left lung apex.}
\end{enumerate}
In the first example, \textit{`scarring'} was annotated as a \textsc{Trajector} after reconciliation as its spatial location (\textit{`left apex'}) is described directly, although there is a higher chance of annotating it as a \textsc{Diagnosis} since most of the probable diagnoses terms are usually preceded by a \textsc{Hedge} term (\textit{`Probably'} in this case). Similarly, \textit{`cavitary lesion'} is indirectly connected to the same body location (\textit{`left apex'}) and has been interpreted as an additional finding. So, \textit{`cavitary lesion'} was also annotated as a \textsc{Trajector} and not as a \textsc{Diagnosis}.
In the second example, \textit{`cavitary lesion'} was annotated as a \textsc{Diagnosis} in context to the first `in' in the sentence, whereas the same term \textit{`cavitary lesion'} was annotated as a \textsc{Trajector} when its role was identified in context to the second `in'. As previously noted, this difference where the same term can be both a \textsc{Trajector} and \textsc{Diagnosis} in different sentences is a consequence of focusing on explicitly representing the spatial language as described as well as the natural ambiguity between a finding and diagnosis in radiology. As a result, some downstream processing or interpretation is still required, which we leave to future work.
 
\subsubsection{Annotation statistics}
A total of 1972 spatial relations are annotated in our corpus of 2000 reports. Most of the \textsc{Trajector} terms were findings, however, 176 out of 2293 terms annotated as \textsc{Trajectors} were anatomical locations (example shown in Figure~\ref{fig:examples}(c-2)). 118 \textsc{Spatial Indicator}s had more than one probable \textsc{Diagnosis}, out of which 98 were associated with 2 \textsc{Diagnosis} terms, 17 were associated with 3 \textsc{Diagnosis} terms, and 3 had 4 associated \textsc{Diagnosis} terms. There are 1062 reports containing at least one sentence triggering a spatial relation. In 1062 reports, there are 1780 sentences each containing at least one \textsc{Spatial Indicator} (1559 sentences containing exactly one \textsc{Spatial Indicator} and remaining 221 containing more than one \textsc{Spatial Indicator}). We have highlighted a brief descriptive statistics of our corpus based on the reconciled version of the annotations in Table~\ref{table:descriptive_stats}.

\begin{table}[t!]
\caption{Annotator agreement.}
	\centering
	\resizebox{\textwidth}{!}{
	\begin{tabular}{lc|cccc}
		\hline 
		\multirow{ 2}{*}{\textbf{Number of Reports}} &
		\multicolumn{1}{c}{\textbf{Kappa (}\textbf{\bm{$\kappa$})}} &
		\multicolumn{4}{c}{\textbf{Overall F1}} \\
		\cline{2-6}
		& \textsc{Spatial Indicator} & \textsc{Trajector} & \textsc{Landmark} & \textsc{Diagnosis} & \textsc{Hedge} \\
		\hline 
		First 500 & 0.88 & 0.44 & 0.50 & 0.25 & 0.49 \\
		\hline
		Remaining 1500 & 0.93 & 0.66 & 0.71 & 0.62 & 0.57 \\
		\hline
		Complete 2000 & 0.92 & 0.59 & 0.64 & 0.49 & 0.55 \\
		\hline
	\end{tabular}
	}
	 \label{table:inter-annotator_agreement}
\end{table}

\begin{table}[ht]
\caption{Descriptive Statistics of the annotations.}
\centering
\begin{tabular}{l|c}
  \hline
 \textbf{Parameter} & \textbf{Frequency} \\ 
  \hline
  Average length of sentence containing spatial relation & 13 \\
 \textsc{Spatial Indicator} & 1972 \\
 \textsc{Trajector} & 2293 \\
 \textsc{Landmark} & 2167 \\
 \textsc{Diagnosis} & 455 \\
 \textsc{Hedge} & 388 \\
  Sentences containing at least 1 \textsc{Spatial Indicator} & 1780 \\
  Maximum number of \textsc{Spatial Indicator} in any sentence & 4 \\
  Spatial relations containing only \textsc{Trajector} and \textsc{Landmark} & 1589 \\
  Spatial relations containing only \textsc{Trajector}, \textsc{Landmark}, and \textsc{Diagnosis} & 9 \\
  Spatial relations containing only \textsc{Trajector}, \textsc{Landmark}, and \textsc{Hedge} & 70 \\
  Spatial relations containing all 4 spatial roles & 304 \\
  Spatial relations containing more than 1 \textsc{Diagnosis} & 118 \\
  Maximum \textsc{Diagnosis} terms associated with any spatial relation & 4 \\
  \hline
\end{tabular}
\label{table:descriptive_stats}
\end{table}


\section{Methods}
\label{sect:methods}
\subsection{Model for spatial relation extraction}
\begin{figure*}[ht]
\includegraphics[width=1.1\textwidth]{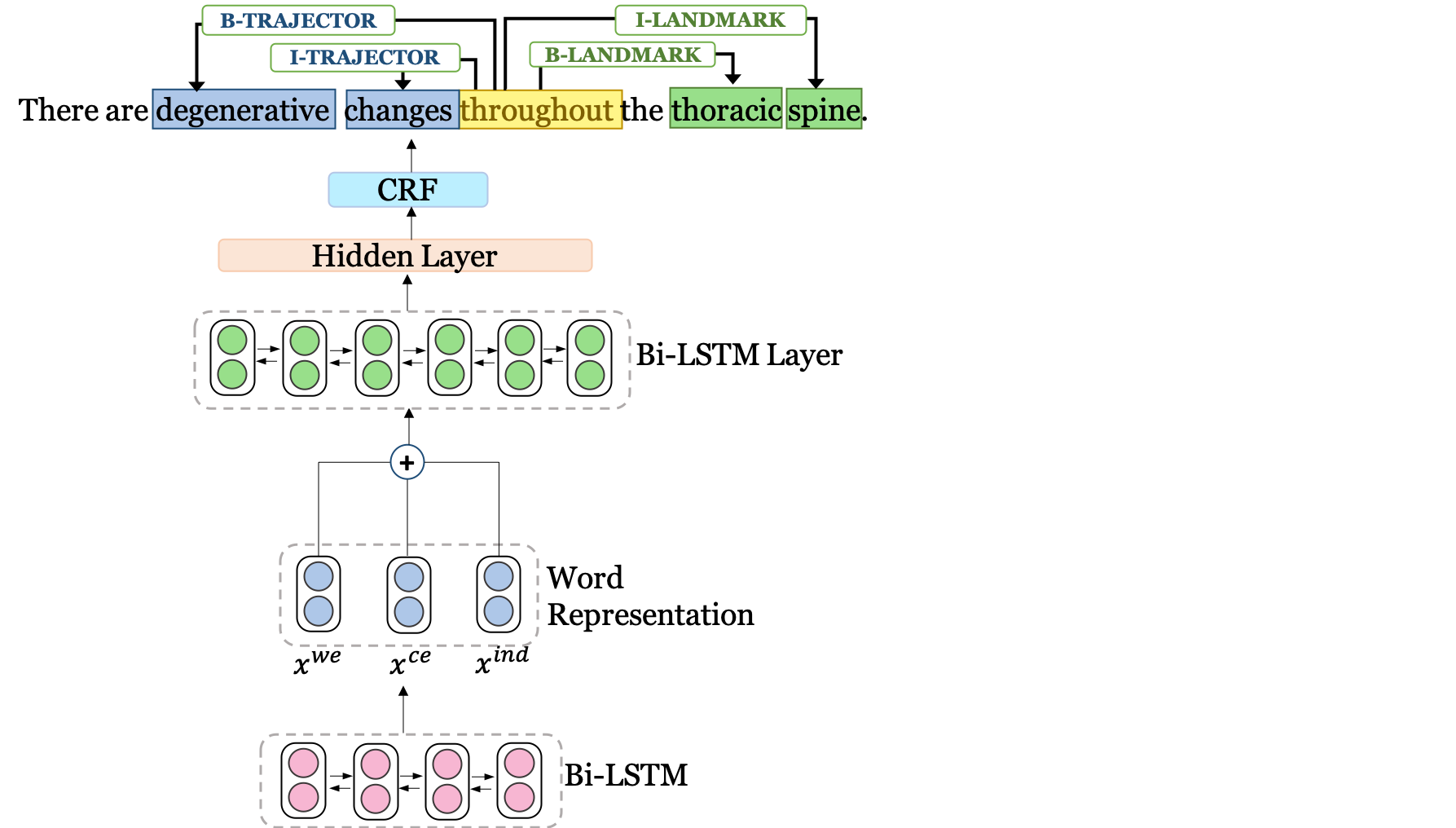}
\centering
\caption{Model architecture. For each word, a character representation is fed into the input layer of the Bi-LSTM network. For each word, $ x^{we} $ represents pre-trained word embeddings, $ x^{ce} $ represents character embeddings, and $ x^{ind} $ represents indicator embeddings. The final predictions for the spatial role labels in a sentence are made combining the Bi-LSTM's final score and CRF score.}
\label{fig:model}
\end{figure*}

We formulate the spatial role extraction as a sequence labeling task. We utilize a Bi-LSTM CRF framework similar to the proposed architecture in Lample et al. \cite{Lample2016neural} for spatial role labeling. The main intention behind using Bi-LSTM is that it works well for taking in the long term dependencies in a sentence, and the bi-directional sequential architecture adds more benefits by considering both the previous and future context of a word. The CRF in the decoding layer takes into account the sequential information in the sentence while predicting the sequence labels related to any spatial role (\textsc{Trajector}/\textsc{Landmark}/ \textsc{Diagnosis}/\textsc{Hedge}). 
We utilize a Bi-LSTM network to incorporate character embedding $ x_i^{ce} $ (where each character is denoted $c_{i,j}$) for each word $ w_i $ in a sentence where $ i $ represents the word position. This is to better handle out-of-vocabulary, rare, and misspelled words. For every word, this character embedding is then concatenated with the respective pre-trained word embedding $ x_i^{we} $. In addition, a {\sc Spatial Indicator} embedding $ x_i^{ind} $
is concatenated to the word and character embeddings to distinguish the indicators from non-indicator words. The final concatenated representation [{$ x_i^{we} $}; $ x_i^{ce} $; $ x_i^{ind} $] is fed into the final Bi-LSTM network with one hidden layer. The overall architecture is presented in Figure~\ref{fig:model}.

\subsection{Pre-processing}
We preprocess the Rad-SpRL dataset to generate input sequence for the model. For each \textsc{Spatial Indicator} in a sentence, we create an instance or sample of the sentence. This results in a total of 1867 sentences with unique position of the \textsc{Spatial Indicator}. For each instance, we tag all the spatial roles (\textsc{Trajector}/\textsc{Landmark}/\textsc{Diagnosis}/\textsc{Hedge}) as well as the \textsc{Spatial Indicator}.
Creating separate sentence instance for each \textsc{Spatial Indicator} helps in dealing with cases where the same word can be both a \textsc{Trajector} and a \textsc{Landmark} in context to two different \textsc{Spatial Indicator}s in the sentence (example shown in (c-1) and (c-2) in Figure~\ref{fig:examples}). Also, annotating only the roles associated with a single \textsc{Spatial Indicator} provides the model unambiguous information about the position of the specific indicator term to which these roles are associated. We follow Beginning (B), Inside (I), and Outside (O) tagging scheme and label the \textsc{Spatial Indicator} term as \textsc{Indicator}. The input to the final Bi-LSTM neural network consists of words and the corresponding B, I, O labels for a set of sentences. 
The following example shows the tagged words for the sentence -- \textit{``Minimal degenerative changes \textbf{of} the thoracic spine''}.
\begin{itemize}
  \itemsep0em
  \item[] $[$\textit{Minimal}]$_{\textsc{B-Trajector}}$ [\textit{degenerative}]$_{\textsc{I-Trajector}}$ [\textit{changes}]$_{\textsc{I-Trajector}}$
  \item[] [\textbf{\textit{of}}]$_{\textsc{Indicator}}$ [\textit{the}]$_{\textsc{O}}$ [\textit{thoracic}]$_{\textsc{B-Landmark}}$ [\textit{spine}]$_{\textsc{I-Landmark}}$
\end{itemize}

\subsection{Experimental settings and evaluation}
10-fold cross validation was performed to evaluate the performance of the Bi-LSTM CRF model in extracting the spatial roles given the gold \textsc{Spatial Indicator} in a sentence. The training, validation, and the test sets consisted of 1495, 186, and 186 sentences (80\%/10\%/10\% splits). We reported the average Precision, Recall, and F1 measures of each of the 4 spatial roles across 10 iterations. We also calculated the overall measures of the three metrics considering all the roles collectively. Exact match was performed for evaluating the model performance on the test set. In addition to the spatial roles, we also evaluated the model performance in extracting the \textsc{Spatial Indicator}s using the same folds.
 
We used pre-trained medical domain MIMIC-III word embeddings of 100 dimensions \footnote{https://northwestern.app.box.com/s/eprxyxmee37p3d6khqbpn125tyttq4u6} in our experiments. The character and the indicator embeddings were initialized randomly and altered during training. The dimensions of character and indicator embeddings were 100 and 5 respectively. The model was implemented using TensorFlow \cite{abadi2016tensorflow}, and the hyperparameters were chosen based on the validation set. LSTM hidden size was set at 500, dropout rate was set at 0.5, learning rate at 0.01, and learning rate decay at 0.99. We used Adam optimizer and trained the model for a maximum of 20 epochs.


\section{Results}
\label{sect:results}
The results of the 10-fold CV for both spatial role extraction and the \textsc{Spatial Indicator} extraction from the Rad-SpRL corpus are presented in Table ~\ref{tab:cv_result}. For spatial role extraction, we considered the gold \textsc{Spatial Indicator} in a sentence. As the results demonstrate, the model performed the best in case of \textsc{Landmark} with an average F1 of $ 94.61 $, followed by \textsc{Trajector} where the average F1 is $ 90.28 $. Using the same Bi-LSTM-CRF architecture, we found that the recall in case of \textsc{Spatial Indicator} extraction is high ($ 99.25 $\%), which suggests that most of the gold indicators were predicted correctly by the system. However, the relative low precision ($ 78.86 $\%) demonstrates that many prepositions in the dataset were wrongly classified as \textsc{Spatial Indicator}, which would consequently generate false positive spatial roles if these predicted indicators are used instead of the gold ones for spatial role labeling.

\begin{table}[ht]
\caption{Average Precision, Recall, and F1 measures of 10-fold CV. \textsc{Sp-In} -- \textsc{Spatial Indicator}.}
    \centering
    \resizebox{\textwidth}{!}{
    \begin{tabular}{m{0.15\linewidth} M{0.1\linewidth} M{0.15\linewidth} M{0.15\linewidth} M{0.15\linewidth} M{0.1\linewidth} M{0.1\linewidth}}
    \hline
    \textbf{Metrics} & \textbf{\textsc{Sp-In}} & \textbf{\textsc{Trajector}} & \textbf{\textsc{Landmark}} & \textbf{\textsc{Diagnosis}} & \textbf{\textsc{Hedge}} & \textbf{\textsc{Overall}} \\
    \hline
Precision (\%) & $ 78.86 $ & $ 90.05 $ &  $ 95.01 $ & $ 70.35 $ &  $ 74.66 $ &   $ 89.21 $ \\
\hline
Recall (\%) & $ 99.25 $ & $ 90.55 $ &  $ 94.23 $ & $ 73.15 $ &  $ 72.67 $ &   $ 89.25 $ \\
\hline
F1 & $ 87.82 $ & $ 90.28 $ & $ 94.61 $ &  $ 71.47 $ &  $ 73.27 $ &  $ 89.22 $ \\
\hline
\end{tabular}
}
\label{tab:cv_result}
\end{table}

\section{Discussion}
\label{sect:discussion}
In this paper, we extract the four spatial roles with respect to a \textsc{Spatial Indicator} in a sentence following the Rad-SpRL annotation scheme. This includes identifying the probable diagnoses with associated hedges in context to a spatial relation between any finding and its associated location. The results in Table ~\ref{tab:cv_result} demonstrate that the Bi-LSTM-CRF model achieves promising results in extracting the spatial roles from the Rad-SpRL corpus. The average F1 measures are around 90 and 95 for extracting \textsc{Trajector} and \textsc{Landmark} respectively. However, the F1 measures are comparatively low for \textsc{Diagnosis} and \textsc{Hedge} roles. The reason behind this can be attributed to the lesser number of \textsc{Diagnosis} and \textsc{Hedge} terms in the dataset (5 to 6 times lesser than both \textsc{Trajector} and \textsc{Landmark} terms) and greater distance between the \textsc{Spatial Indicator} and the \textsc{Diagnosis}/\textsc{Hedge} terms compared to the \textsc{Trajector}/\textsc{Landmark} terms.

Taking into account the relatively low F1 measure for \textsc{Diagnosis}, we analyzed the errors and many of these are due to the challenging nature of the description used by radiologists to express their interpretation. For example, in the sentence -- \textit{``Low lung volumes \textbf{with} bibasilar opacities may represent bronchovascular crowding.''}, the \textsc{Diagnosis} \textit{`bronchovascular crowding'} is falsely classified as a \textsc{Trajector}. This might be because there are many instances in the dataset where \textit{`bronchovascular crowding'} appears as \textsc{Trajector}, as often a \textsc{Diagnosis} term itself appears in a spatial relationship.

In this work, we have considered both positive and negative spatial relations. We aim to differentiate the negated relations in future. Future work should also be directed toward building an end-to-end system based on neural joint learning models \cite{Li2017neuraljoint, Miwa2016end-to-end} that would extract both \textsc{Spatial Indicator} and the spatial roles together. We further aim to consider non-prepositional spatial expressions as \textsc{Spatial Indicator}s (e.g., verbs such as \textit{`demonstrates'}, \textit{`shows'} etc.) that indicate the presence of any spatial relation between finding and body location. Besides clinical findings, we also intend to extend the Rad-SpRL framework to extract other important and common spatially-grounded radiology entities such as medical devices from the reports. Following example illustrates a sample sentence where the medical device \textit{`Right IJ venous catheter'} acts as the \textsc{Trajector} in context to its associated location \textit{`proximal SVC'} that acts as the \textsc{Landmark}:
\begin{itemize}
    \item[]\textit{Right IJ venous catheter terminates \textbf{at} the proximal SVC.}
\end{itemize}
Another limitation of this work is that the inter-sentence spatial relations are not covered, although the frequency of such cases are very rare in the Rad-SpRL corpus. We also aim to evaluate the generalizability of our sequence labeling method in extracting the spatial roles from datasets across institutions. For standardization of the extracted spatial roles, we further aim to normalize them utilizing the existing radiology lexicons such as RadLex \cite{langlotz2006radlex} codes. From a method perspective, we plan to apply more recent deep learning methods such as BERT-based models \cite{devlin2019bert}, highway networks \cite{srivastava2015highway}, and syntax-enhanced models \cite{He2018syntaxforSRL} to improve the existing performance of spatial role extraction from the Rad-SpRL corpus. 

\section{Conclusion}
\label{sect:conclusion}
This paper proposes a spatial representation framework in radiology (Rad-SpRL). It provides a detailed description of the annotation scheme used for extracting spatial information from radiology reports. This consists of annotating four radiology-specific spatial roles in a dataset of 2000 chest X-ray reports. The spatial roles are annotated in the context of a \textsc{Spatial Indicator} which denotes the presence of a spatial relation between clinical findings and body locations. It additionally identifies probable diagnoses and hedging terms associated with the spatially-related \textit{finding}-\textit{location}. For this, we employ a Bi-LSTM-CRF model to automatically extract the spatial roles from our annotated Rad-SpRL corpus. The model achieves satisfactory performance with average F1 measures of around 90, 95, 71, and 73 for \textsc{Trajector}, \textsc{Landmark}, \textsc{Diagnosis}, and \textsc{Hedge} roles, respectively. In the future, we aim to evaluate the model on much larger datasets, improve the current performance by using more recent and advanced deep learning techniques, and develop joint learning models for extracting the \textsc{Spatial Indicator} and the spatial roles jointly.


\bibliography{mybibfile}

\end{document}